\documentclass{article}

\usepackage[eandd,preprint]{neurips_2026}

\usepackage[utf8]{inputenc}
\usepackage[T1]{fontenc}

\usepackage{microtype}
\usepackage{xspace}  

\usepackage{amsfonts}
\usepackage{amsmath}
\usepackage{nicefrac}

\usepackage{booktabs}
\usepackage{multirow}
\usepackage{threeparttable}

\usepackage{graphicx}
\usepackage{subcaption}
\usepackage{float}  

\usepackage{enumitem}

\usepackage[table]{xcolor}
\usepackage{hyperref}
\hypersetup{
  colorlinks=true,
  linkcolor=blue!60!black,
  citecolor=green!50!black,
  urlcolor=blue!70!black,
}

\usepackage{tikz}
\usepackage{pgfplots}
\pgfplotsset{compat=1.18}
\usetikzlibrary{shapes,arrows,positioning,calc}

\usepackage{algorithm}
\usepackage{algorithmic}

\newcommand{\eg}{e.g.\@\xspace}

\usepackage{tabularx}

\usepackage{listings}
\usepackage[listings]{tcolorbox}
\newtcblisting{promptbox}[2][]{
    colback=gray!5,
    colframe=gray!60!black,
    arc=2pt, outer arc=2pt,
    boxrule=0.8pt,
    listing only,
    listing options={
        basicstyle=\ttfamily\small,
        breaklines=true,
        breakatwhitespace=true,
        columns=fullflexible,
        keepspaces=true,
        language=,
        aboveskip=0pt,
        belowskip=0pt,
        showstringspaces=false
    },
    title={\textbf{#2}},
    #1
}

\makeatletter
\renewcommand{\@notice}{}
\makeatother

\title{DiffSpot: Can VLMs Spot Fine-Grained Visual Differences in Web Interfaces?}

\author{
    \textbf{Linhao Zhang\thanks{Correspondence to \texttt{zhanglinhao90@gmail.com}.}}\textmd{,}
    \textbf{Aiwei Liu}\textmd{,}
    \textbf{Yuan Liu}\textmd{,}
    \textbf{Xiao Zhou}
    \\
    \\
    WeChat AI, Tencent Inc.
}

\begin{document}

\maketitle

\begin{abstract}
Vision-language models (VLMs) have made strong progress on high-level
image-text alignment, yet their ability to perceive subtle visual differences
remains limited. We study this problem in rendered web interfaces, where
localized visual changes are both a diagnostic test of fine-grained perception
and a practical requirement for GUI agents and design tools. We introduce
\textbf{DiffSpot}, a code-driven benchmark for open-ended spot-the-difference
on web interfaces. DiffSpot constructs controlled image pairs by mutating a
single CSS property of a target element in self-contained HTML, re-rendering
the page, and recording the changed property, element, and mutation magnitude.
A grounding gate retains only pairs whose rendered pixel difference is confined
to the target element. The benchmark contains 4{,}400 pairs, including
3{,}900 has-diff pairs balanced across 13 CSS-property operators and three
difficulty tiers, plus 500 no-diff pairs for hallucination control.
Evaluating 13 frontier VLMs zero-shot, we find that even the best model
identifies only $40.7\%$ of true changes, with Hard-tier Recall below
$23\%$ for every model. DiffSpot further shows that difficulty is strongly property-dependent:
across CSS operators, neither pixel magnitude nor CLIP distance reliably
predicts Recall.

\end{abstract}

\begin{center}
\footnotesize
\begin{tabular}{rcl}
\raisebox{-1.5pt}{\includegraphics[height=1.05em]{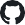}} & \textbf{Code} & \href{https://github.com/Tencent/DiffSpot}{\nolinkurl{https://github.com/Tencent/DiffSpot}} \\
\raisebox{-1.5pt}{\includegraphics[height=1.05em]{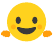}} & \textbf{Data} & \href{https://huggingface.co/datasets/tencent/DiffSpot}{\nolinkurl{https://huggingface.co/datasets/tencent/DiffSpot}} \\
\end{tabular}
\end{center}

\section{Introduction}
\label{sec:introduction}

\begin{figure}[t]
\centering
\includegraphics[width=\textwidth]{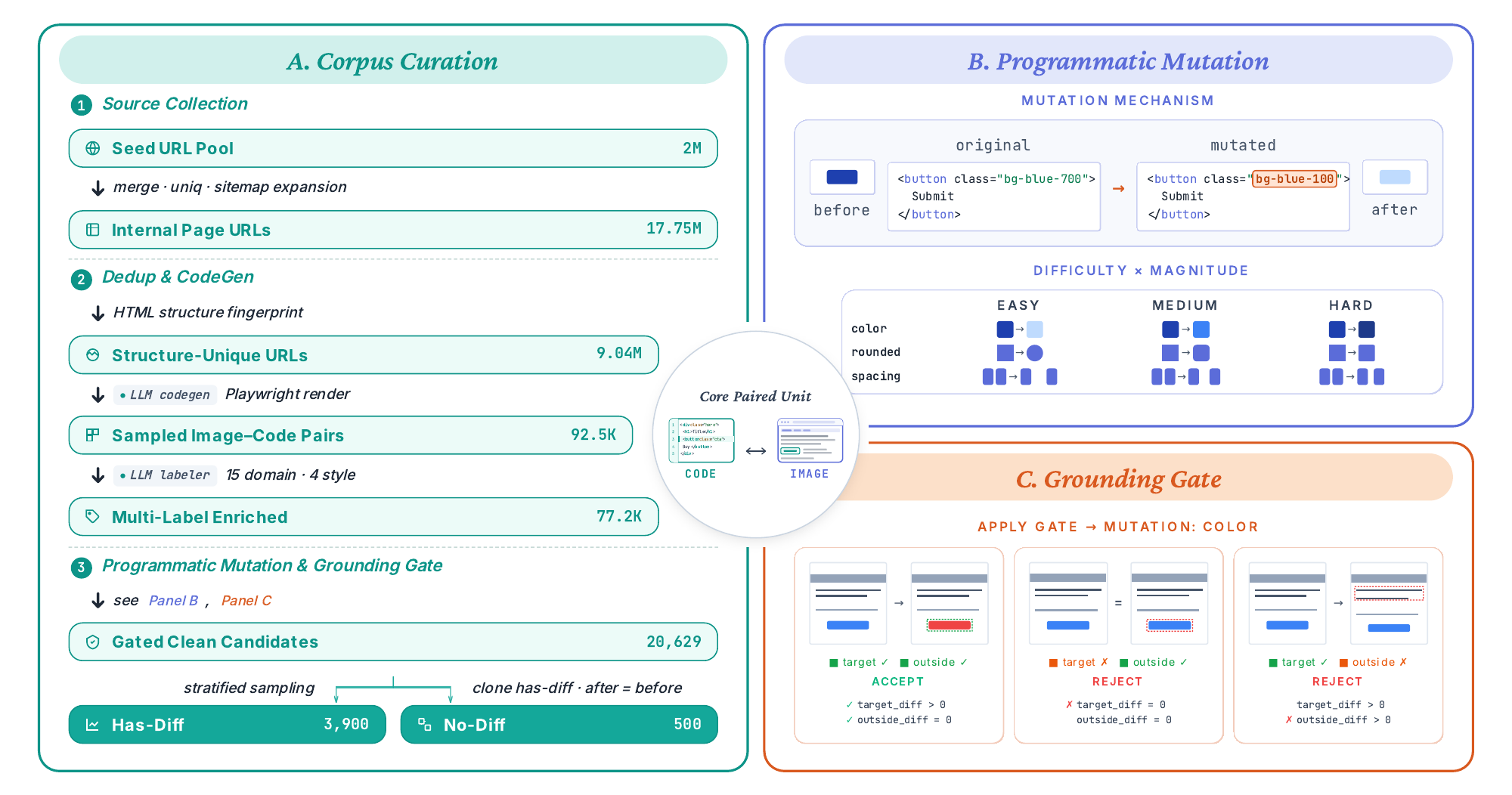}
\caption{\textbf{DiffSpot construction pipeline.}
DiffSpot turns real web pages into controlled before/after screenshot pairs
by moving visual-difference construction from image space to code space.
\textbf{A.}~Corpus curation collects and filters rendered web-interface
candidates from URL-seeded pages.
\textbf{B.}~Programmatic mutation applies single CSS-property changes across
operator-specific difficulty tiers.
\textbf{C.}~A grounding gate validates the rendered result, retaining only
pairs whose pixel difference is confined to the target element.}
\label{fig:pipeline}
\end{figure}

Vision-language models (VLMs) have made strong progress on high-level
image-text alignment and language-guided visual reasoning, yet remain brittle
on fine-grained visual perception~\citep{tong2024eyes,luo2025vilp,shahgir2026vlmswords}.
A direct stress test of \emph{fine-grained perception} is
\emph{spot-the-difference}: given two nearly identical screenshots, can a
model tell exactly what changed? In such near-identical pairs, semantic
shortcuts provide little help; the model must localize and name a small
visual change against an otherwise unchanged background. This capability is
not only diagnostic, but also practically necessary for systems that operate
on rendered web interfaces, including GUI agents and design
tools~\citep{anthropic2025claude}. Such systems must verify not merely that
the screen changed after an action or edit, but which visual property
changed, where it changed, and whether the change was confined to the
intended element.

Despite this diagnostic and practical need, web-interface
spot-the-difference remains largely absent from existing pair-difference
benchmarks for VLMs~\citep{omnidiff,vlmsubtlebench}. The bottleneck is
twofold. First, it is difficult to collect the right image pairs at scale:
real web pages rarely provide near-identical before/after screenshots that
differ by only one small UI property, let alone balanced coverage over UI
elements, visual properties, and difficulty levels. Second, even when similar
pairs are available, post-hoc human labeling is a selective filter rather than
a neutral sampler of visual differences. Human change detection is known to be
attention-, semantics-, and change-type dependent, rather than determined by
pixel magnitude alone~\citep{rensink1997see,hollingworth2000semantic,cole2006change,wright2005saliency,stirk2007low}.
As a result, annotation tends to over-represent changes that are visually
salient, semantically meaningful, or easy to name, while subtle
visual-property changes are more likely to be missed, labeled at inconsistent
granularity, or unevenly covered across properties and difficulty levels.
Thus, the target regime for web-interface diffing---subtle, localized,
property-level differences---remains under-covered.

Our key insight is to move difference construction from image space to code
space. Unlike natural images, web interfaces are programmatic visual artifacts:
their pixels are generated by rendering structured HTML/CSS. This lets us
define a visual difference before an image pair is created, rather than
discover one after the fact. Given a self-contained HTML page, we mutate a
single CSS property of a target element and re-render the page in a headless
browser. The resulting before/after screenshots are paired with a
machine-readable mutation record specifying the changed property, target
element, and mutation magnitude. By varying the mutation magnitude within the
same CSS operator, we can parameterize difficulty while holding the visual
property fixed. This code-driven construction directly addresses both
bottlenecks: it systematically generates near-identical pairs with balanced
coverage across UI elements, visual properties, and difficulty levels, while
avoiding post-hoc annotation bias by specifying each difference before
rendering.

However, specifying differences in code does not guarantee that they appear as
clean, localized differences after rendering. A mutation can be ineffective,
when the CSS change is shadowed and produces no visible effect, or non-local,
when it triggers layout reflow and changes regions beyond the target element.
In either case, the mutation record no longer corresponds to a single localized
visual effect in the rendered screenshots. To enforce this correspondence, we
introduce a \textbf{grounding-gate mechanism} that validates each candidate
mutation in rendered pixel space. Using the target element specified in the
mutation record, we obtain its browser-rendered bounding box and accept a pair
only when the rendered pixel difference is confined to this target region. The
retained pairs therefore align each machine-readable mutation record with
exactly one effective, localized visual difference at the intended element.

We instantiate this idea in \textbf{DiffSpot}, a construction pipeline that
turns real web pages into controlled before/after screenshot pairs. Starting
from URLs collected from 2M source domains, we crawl and render pages, convert
them into self-contained HTML sources, and apply CSS-property mutations
followed by grounding-gate filtering. The resulting benchmark contains 4{,}400
image pairs: 3{,}900 has-diff pairs balanced across 13 CSS-property operators
and three difficulty tiers (100 pairs per operator--tier cell), plus 500
no-diff pairs for hallucination control.

A zero-shot evaluation of 13 frontier
VLMs~\citep{openai2025gpt5,anthropic2025claude,google2025gemini,bai2025qwen,team2026kimi25,team2025thinking,wang2025internvl}
shows that fine-grained visual difference detection on web interfaces remains
far from solved: even the best model identifies only $\mathbf{40.7\%}$ of true
visual changes, and Hard-tier Recall stays below $23\%$ for every model.
Beyond this low ceiling, DiffSpot reveals a property-level failure pattern.
Neither bbox-level pixel change nor CLIP image distance reliably predicts
Recall, suggesting that models struggle to perceive and name CSS-level visual
properties rather than simply detect larger image differences. Moreover,
performance varies much more across CSS operators than across source domains:
what changed matters more than where it appeared. The 500 no-diff pairs
further expose a sensitivity--restraint trade-off, disentangling sensitivity
to real changes from restraint on unchanged pairs.

Our contributions are:

\begin{itemize}[leftmargin=*,itemsep=2pt,topsep=2pt]
\item \textbf{A web-interface spot-the-difference benchmark.}
We introduce \textbf{DiffSpot}, to our knowledge the first benchmark for
open-ended spot-the-difference on rendered web interfaces.

\item \textbf{A code-driven visual-difference generation pipeline.}
We develop a pipeline that creates controlled visual differences by
programmatically mutating CSS properties in self-contained HTML and validating
the rendered result with a bounding-box grounding gate.

\item \textbf{Property-level diagnostic findings.}
We evaluate 13 frontier VLMs zero-shot and reveal property-specific failures:
even the best model identifies only $40.7\%$ of true visual changes, while
pixel- and CLIP-based magnitudes poorly predict Recall.
\end{itemize}

\section{Benchmark Construction}
\label{sec:construction}

DiffSpot is built by a five-stage pipeline (Figure~\ref{fig:pipeline})
that turns a large pool of rendered web pages into a quality-gated benchmark
balanced across operator--difficulty cells: source corpus curation
(\S\ref{sec:source}), programmatic mutation (\S\ref{sec:mutation}),
grounding-gate validation (\S\ref{sec:gate}), polish and filtering
(\S\ref{sec:filter}), and stratified sampling with a no-diff control
(\S\ref{sec:sampling}). All ground truth is derived programmatically from the
structured mutation record.

\subsection{Source Corpus Curation}
\label{sec:source}

We seed the corpus from the Chrome User Experience Report Top-1M
and Majestic Top-1M (2M domains; 1.35M after host dedup), expand
each domain via its sitemap to 17.75M page URLs, and keep one URL
per HTML-structure fingerprint (9.04M structure-unique URLs). Each
URL is rendered in headless Chromium driven by
Playwright~\citep{playwright} at a fixed $1280\times800$ viewport
and paired with a self-contained HTML produced by LLM regeneration,
keeping the released dataset free of third-party licensed content.
We retain only pairs whose CLIP~\citep{clip} similarity between the
original and regenerated renders is $\geq 0.70$, yielding the sampled
image--code pair pool shown in Figure~\ref{fig:pipeline}. A three-VLM-judge
realness audit on 501 pairs scores renderings within 0.3 points of originals
on a 5-point scale (\S\ref{sec:realness_audit}). After rule-based content
filtering for PII, abnormal HTML body length, and dynamic tags, an LLM
domain/style labeler (\texttt{gpt-oss-120b}~\citep{openai2025model}) and a
capped sampling policy yield the \emph{multi-label enriched pool} that
feeds the mutation stage (Figure~\ref{fig:pipeline};
\S\ref{app:construction-details}).

\subsection{Programmatic Mutation}
\label{sec:mutation}

DiffSpot is \emph{by design} restricted to atomic, localized visual
differences: each pair isolates a single CSS-property-level mutation on a
single target element. This scope is essential for a per-property capability
probe. Compound or reflow-heavy changes can introduce coarse visual cues and
break the correspondence between the mutation record and the rendered
screenshots; the resulting pair no longer isolates which CSS-level visual
property a model actually perceives and names.

We define 13 CSS-property-level operators grouped into four
families: \emph{typography} (\texttt{font\_weight},
\texttt{font\_size}, \texttt{letter\_spacing},
\texttt{line\_height}, \texttt{text}), \emph{color}
(\texttt{color}, \texttt{opacity}, \texttt{gradient}),
\emph{layout} (\texttt{position}, \texttt{spacing},
\texttt{justify}), and \emph{shape} (\texttt{border},
\texttt{rounded}). Two mutation mechanisms are selected per
operator---a Tailwind-CSS~\citep{tailwindcss} class swap and an
inline-style override; both operate on the
static HTML and are fully reproducible
(\S\ref{app:mutation-mechanics}).

\paragraph{Difficulty tiers.}
Each operator is stratified into Easy, Medium, and Hard tiers with
non-overlapping parameter ranges, so the difficulty tier of a
candidate is parameterized solely by the magnitude of the property
change. For step-based operators (\eg \texttt{rounded},
\texttt{color}), tiers correspond to Tailwind-scale step distance
(Easy: 3--5, Medium: 2, Hard: 1); for continuous-valued operators
(\eg \texttt{letter\_spacing}), tiers are em-offset magnitude
(Easy: $\pm0.20$\,em, Medium: $\pm0.12$\,em, Hard:
$\pm0.06$\,em). Full parameter ranges are in
Table~\ref{tab:op-principles} (\S\ref{app:operator-rules}).
Difficulty is strictly ordered \emph{within} each operator. Each pair is
processed in grouped mode (one candidate per tier) to produce the raw
mutation candidate pool, which the grounding gate then validates
(Figure~\ref{fig:pipeline}, Panel C).

\begin{figure}[t]
  \centering
  \includegraphics[width=0.9\textwidth]{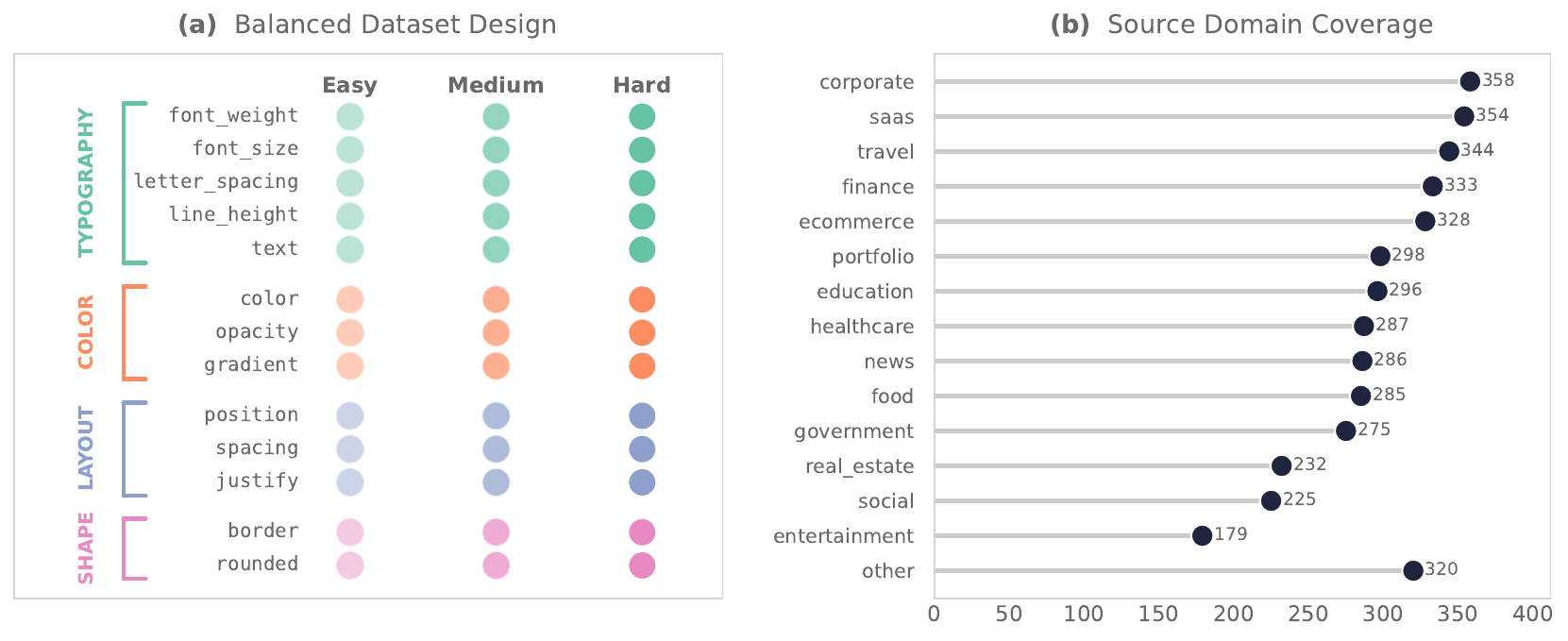}
  \caption{\textbf{DiffSpot dataset statistics.}
  \textbf{(a)~Balanced design.} 13 operators $\times$ 3 difficulty
  tiers $=$ 39 cells with 100 has-diff pairs each (3{,}900 total),
  plus 500 no-diff pairs. Color encodes operator family; shade
  encodes difficulty.
  \textbf{(b)~Source-domain coverage.} All 15 domain categories,
  sorted by frequency.}
  \label{fig:stats}
\end{figure}

\subsection{Grounding Gate}
\label{sec:gate}

Naively re-rendering a mutated HTML can break the intended
code-to-pixel correspondence in two ways: \emph{no-effect mutations},
where the CSS override is silently shadowed and produces no visible
effect, and \emph{reflow contamination}, where the mutation cascades
beyond the target element and changes other regions of the page. A
full-image pixel-diff filter can verify that the page changed, but
not that the change is localized to the intended element. We therefore
anchor validation to the target element's bounding box, queried
declaratively from the rendered
DOM; the bbox is never inferred from pixel content. The gate requires
three predicates:
\begin{enumerate}[leftmargin=*,itemsep=0pt,topsep=0pt]
\item \textbf{Effectiveness.} Inside-bbox pixel change is non-zero.
\item \textbf{Locality.} Outside-bbox region is unchanged at pixel level.
\item \textbf{Selector resolution.} Selectors that fail to resolve in the
rendered DOM are rejected.
\end{enumerate}
These predicates jointly retain only pairs in which the rendered pixel
change is concentrated inside the target bounding box and absent
outside it.

\subsection{Polish and Filtering}
\label{sec:filter}

We polish each gated record into a natural-language answer using
\texttt{gpt-oss-120b}~\citep{openai2025model} at temperature $0.7$
with paraphrase exemplars. Each pair ships with both the structured
mutation record, which defines the scoring ground truth, and the
natural-language description, which is used only for display. A small
set of content filters then removes residual quality-failure cases
(\S\ref{app:construction-details}).

\subsection{Stratified Sampling and No-Diff Control}
\label{sec:sampling}

We partition the filtered candidates into 39 cells (13 operators
$\times$ 3 difficulty tiers) and draw exactly 100 candidates per
cell (3{,}900 total). The $n{=}100$ choice gives a per-cell
binomial standard error of $\approx 5$\,pp at $p{=}0.5$, tight
enough for per-cell reporting in \S\ref{sec:property_failures}. To
measure hallucination, we add 500 no-diff pairs constructed by
rendering the same HTML twice (no mutation applied); the ground-truth
answer is ``No visible differences.''

\subsection{Final Composition}
\label{sec:stats}

The final DiffSpot benchmark contains \textbf{4{,}400 image pairs}:
3{,}900 has-diff pairs (100 per operator-tier cell, 39 cells) and
500 no-diff pairs. Figure~\ref{fig:stats} summarises the balanced
design and source-domain coverage. Each pair ships with the
before/after PNG screenshots, the structured mutation record, and
the polished natural-language description; the benchmark is fully
regenerable from the released self-contained HTML under a
deterministic rendering pipeline.

\section{Experiments}
\label{sec:experiments}

\subsection{Setup}

\label{sec:exp_setup}

\paragraph{Models.}
We evaluate 13 recent and frontier vision-language models on DiffSpot zero-shot across
proprietary API and open-weight access categories.
\textbf{Proprietary API:} four models accessible only through vendor
APIs~\citep{google2025gemini,google2026gemini31pro,google2025gemini3flash,openai2025gpt5,openai2025gpt54,anthropic2025claude}.
\textbf{Open-weight:} nine models with publicly released weights:
Kimi~K2.5~\citep{team2026kimi25},
Qwen3.5-VL-397B~\citep{qwen2026qwen35vl},
GLM-4.6V/-Flash~\citep{team2025thinking}, a 2$\times$2 Qwen3-VL grid
(\{30B, 235B\} $\times$ \{Instruct, Thinking\})~\citep{bai2025qwen},
and InternVL3.5-30B-A3B~\citep{wang2025internvl}. Full names and
parameter counts are in Table~\ref{tab:main_table}; the 2$\times$2
Qwen3-VL grid lets us isolate the effect of reasoning mode at two scales.

\paragraph{Inference.}
All models use greedy decoding (temperature $0$) with a 16{,}384-token
output budget; image pairs are fed at the original $1280\times800$ viewport
resolution (\S\ref{sec:construction}). The prompt is a single zero-shot
instruction that presents the before/after screenshots and asks the model to
list observed differences, without worked examples or hints. The prompt is
identical across models.

\paragraph{Metrics.}
We reduce each open-ended response to a per-case binary verdict and report
\textbf{Accuracy} $= (TP+TN)/4{,}400$, where $TP$ counts has-diff pairs whose
ground-truth mutation is identified and $TN$ counts no-diff pairs correctly
reported as having no change. Sliced views break this score down
into \textbf{Easy~/~Med~/~Hard~Recall} (correct identifications among the
1{,}300 has-diff cases in each tier) and \textbf{No-Diff Acc.}
($= 1-\text{hallucination rate}$ on the 500 no-diff pairs). Matching is
performed by \texttt{gpt-oss-120b}~\citep{openai2025model} under a
visual-effect-equivalence rubric that is tolerant to paraphrases; for example,
``thicker text'' credits a \texttt{font\_weight}~$400\!\to\!700$ mutation.
The three judge LLMs (\texttt{gpt-oss-120b}, \texttt{Kimi K2.5},
\texttt{Qwen3.5-VL-397B}) reach mean pairwise Cohen $\kappa = 0.93$ and produce
identical 13-model rankings (\S\ref{sec:judge_robustness}); the full prompt is
in \S\ref{app:judge-prompt}.

\subsection{Main Results}
\begin{table}[t]
\centering
\caption{\textbf{Visual Diff Detection on DiffSpot} (percentages,
4{,}400 pairs). \textbf{Easy/Med/Hard} are Recall on the 1{,}300
has-diff cases per tier and \textbf{Diff Overall} is Recall on
3{,}900 has-diff pairs. \textbf{No-Diff} is specificity on 500
no-diff pairs. The shaded \textbf{Overall} column is per-case
Accuracy $(TP+TN)/4{,}400$ and is the leaderboard score. Params:
total\,/\,active for MoE; total only for dense; ``---'' for
proprietary. Rows are sorted by Overall within each access category.
\textbf{Bold}: column max; \underline{underline}: Overall runner-up.}
\label{tab:main_table}
\footnotesize
\setlength{\aboverulesep}{0pt}
\setlength{\belowrulesep}{0pt}
\renewcommand{\arraystretch}{1.25}
\setlength{\tabcolsep}{6pt}
\definecolor{WeBlue}{HTML}{E8F2FA}
\definecolor{RowGray}{HTML}{EEEEEE}
\begin{tabularx}{\textwidth}{@{} >{\raggedright\arraybackslash}X c | rrr r | r >{\columncolor{WeBlue}}r }
\toprule
\multirow{2}{*}{\textbf{Model}} & \multirow{2}{*}{\textbf{Params}} & \multicolumn{4}{c|}{\textbf{Diff}} & \multirow{2}{*}{\textbf{No-Diff}} & \cellcolor{WeBlue} \\
\cmidrule{3-6}
 &  & \textbf{Easy} & \textbf{Med} & \textbf{Hard} & \textbf{Overall} & & \multirow{-2}{*}{\cellcolor{WeBlue}\textbf{Overall}} \\
\midrule
\rowcolor{RowGray} \multicolumn{8}{c}{\textbf{\textit{Open-weight models}}} \\
\midrule
Kimi~K2.5               & 1T\,/\,32B   & 54.2          & 36.4          & 18.6          & \underline{36.4} & 87.2           & \underline{42.2} \\
Qwen3.5-VL-397B         & 397B\,/\,17B & 45.1          & 31.5          & 13.7          & 30.1             & 96.6           & 37.6 \\
Qwen3-VL-235B-Thinking  & 235B\,/\,22B & 30.1          & 17.3          & 10.5          & 19.3             & 98.8           & 28.3 \\
GLM-4.6V-Flash          & 9B           & 24.5          & 17.6          & 9.3           & 17.1             & 75.8           & 23.8 \\
GLM-4.6V                & 106B\,/\,12B & 17.0          & 10.9          & 5.5           & 11.2             & 99.6           & 21.2 \\
Qwen3-VL-30B-Instruct   & 30B\,/\,3B   & 14.5          & 9.0           & 4.5           & 9.3              & 82.0           & 17.6 \\
Qwen3-VL-30B-Thinking   & 30B\,/\,3B   & 16.5          & 8.8           & 3.8           & 9.7              & 77.8           & 17.5 \\
Qwen3-VL-235B-Instruct  & 235B\,/\,22B & 9.6           & 3.0           & 2.6           & 5.1              & \textbf{100.0} & 15.9 \\
InternVL3.5-30B-A3B     & 30B\,/\,3B   & 4.7           & 3.9           & 3.8           & 4.2              & \textbf{100.0} & 15.0 \\

\midrule
\rowcolor{RowGray} \multicolumn{8}{c}{\textbf{\textit{Proprietary models}}} \\
\midrule
Gemini~3.1~Pro          & ---          & \textbf{60.5} & \textbf{38.9} & \textbf{22.7} & \textbf{40.7}    & 98.4           & \textbf{47.2} \\
Gemini~3~Flash          & ---          & 52.5          & 32.5          & 18.2          & 34.4             & 91.4           & 40.9 \\
Claude~Opus~4.7         & ---          & 41.2          & 30.5          & 21.8          & 31.2             & 99.6           & 38.9 \\
GPT-5.4                 & ---          & 48.8          & 30.5          & 12.2          & 30.5             & 99.6           & 38.3 \\
\bottomrule
\end{tabularx}
\end{table}

\label{sec:main_results}

Table~\ref{tab:main_table} reports Visual Diff Detection performance
for all 13 models on the full 4{,}400-pair benchmark.

\textbf{A low ceiling on true-change detection.}
Gemini~3.1~Pro leads the leaderboard at 47.2\% Accuracy, 5.0\,pp ahead of
Kimi~K2.5 (42.2\%) and 32.2\,pp above InternVL3.5 (15.0\%). Yet this
aggregate score masks a sharper failure on the has-diff slice: even the leader
identifies only 40.7\% of ground-truth mutations, missing roughly three of
every five true visual changes. Seven of thirteen models fall below 30\%
Accuracy, and two models (Qwen3-VL-235B-Instruct and InternVL3.5) clear the
trivial always-no-diff baseline of 11.4\% Accuracy by less than 5\,pp.
Open-ended Visual Diff Detection on real web-UI pairs is therefore far from
solved.

\textbf{Hard mutations remain difficult for every model.}
Recall drops sharply from Easy to Hard across the strongest non-abstaining
models: Gemini~3.1~Pro falls from 60.5\% to 22.7\% ($-$37.8\,pp), GPT-5.4 from
48.8\% to 12.2\% ($-$36.6\,pp), Kimi~K2.5 from 54.2\% to 18.6\% ($-$35.6\,pp),
and Gemini~3~Flash from 52.5\% to 18.2\% ($-$34.3\,pp). Hard-tier Recall stays
below 23\% for every model, indicating that the hardest cells expose a
substantially sharper perception failure rather than merely a weaker version
of the Easy setting.

\textbf{No-diff pairs separate sensitivity from restraint.}
The no-diff slice reveals that higher Recall can come with hallucinated
differences. Aggressive reporters hallucinate frequently: GLM-4.6V-Flash marks
24.2\% of no-diff pairs as changed, and the Qwen3-VL 30B variants hallucinate
on 18--22\% of no-diff pairs, reducing their Accuracy relative to has-diff
Recall. At the other extreme, Claude~Opus~4.7, GPT-5.4, and GLM-4.6V
hallucinate on only 0.4\% of no-diff pairs; Qwen3-VL-235B-Instruct and
InternVL3.5 reach 100.0\% No-Diff specificity largely by reporting almost
nothing on either changed or unchanged inputs. Thus, no-diff controls are
necessary to distinguish genuine visual sensitivity from either over-reporting
or abstention.

\textbf{Reasoning helps only at larger scale.}
The four Qwen3-VL variants form a 2$\times$2 grid over \{30B, 235B\}
$\times$ \{Instruct, Thinking\}. At 30B, switching to Thinking leaves Accuracy
essentially unchanged ($17.6\,\to\,17.5$, $-$0.1\,pp). At 235B, the same switch
improves Accuracy by 12.4\,pp ($15.9\,\to\,28.3$), driven mainly by a 20.5\,pp
gain in Easy Recall ($9.6\,\to\,30.1$). Within the Instruct setting, scaling
from 30B to 235B does not improve performance ($17.6\,\to\,15.9$), suggesting
that size alone is insufficient; in this model family, reasoning mode becomes
useful only at larger scale.

\paragraph{Sensitivity--restraint trade-off.}
Plotting has-diff Recall against no-diff hallucination rate
(Appendix~\ref{app:tradeoff}) reveals a Pareto frontier reached only by the
three closed-source frontier APIs: Gemini~3.1~Pro, Claude~Opus~4.7, and
GPT-5.4. No open-weight model enters the ``accurate and restrained'' region
(Recall $\geq30\%$, hallucination $\leq5\%$), showing that current models must
jointly improve sensitivity to real changes and restraint on unchanged pairs.

\subsection{Property-Level Failure Modes}
\begin{figure}[t]
  \centering
  \begin{subfigure}{0.49\textwidth}
    \centering
    \includegraphics[width=\textwidth]{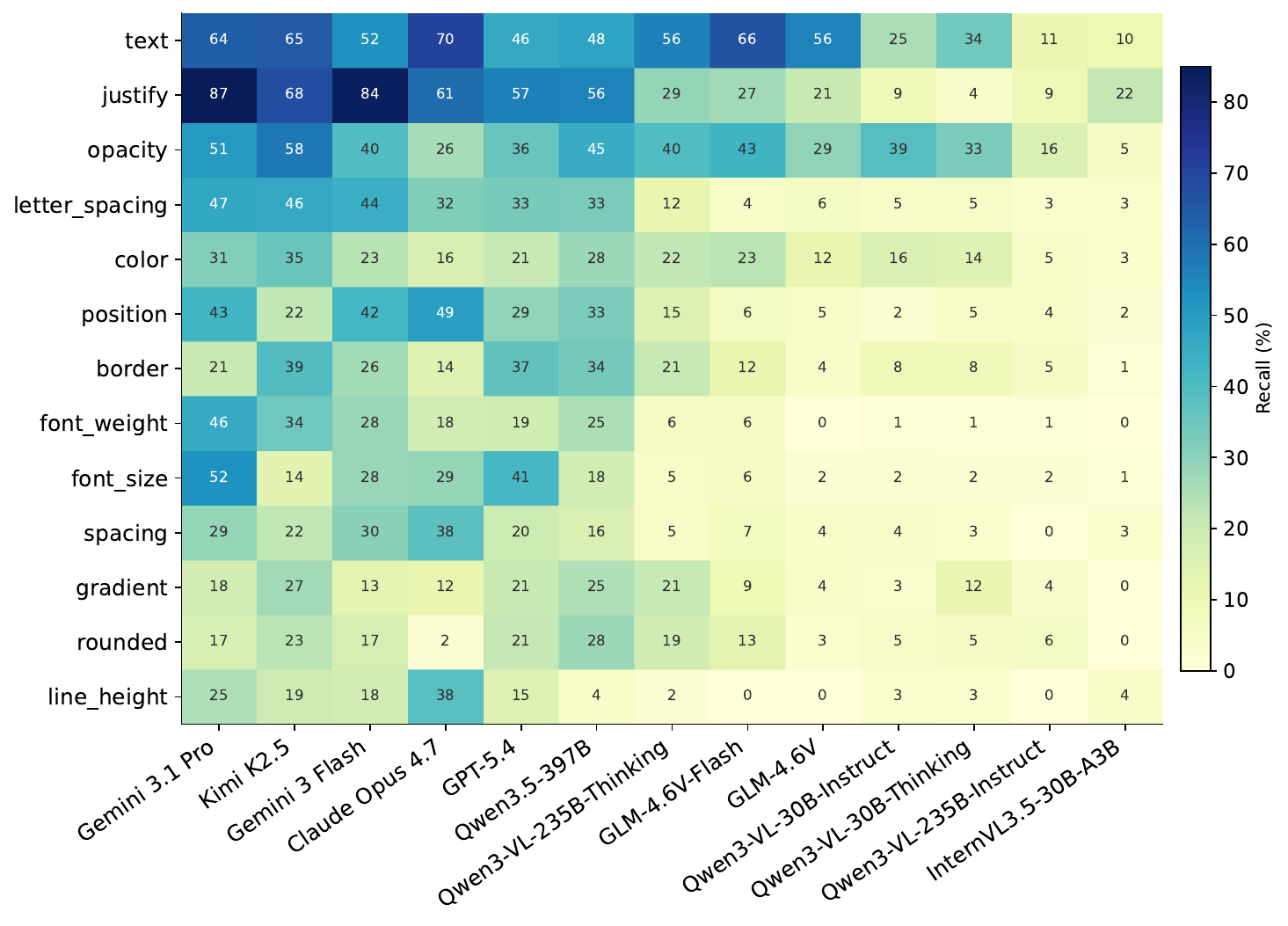}
    \phantomcaption
    \label{fig:heatmap}
  \end{subfigure}
  \hfill
  \begin{subfigure}{0.49\textwidth}
    \centering
    \includegraphics[width=\textwidth]{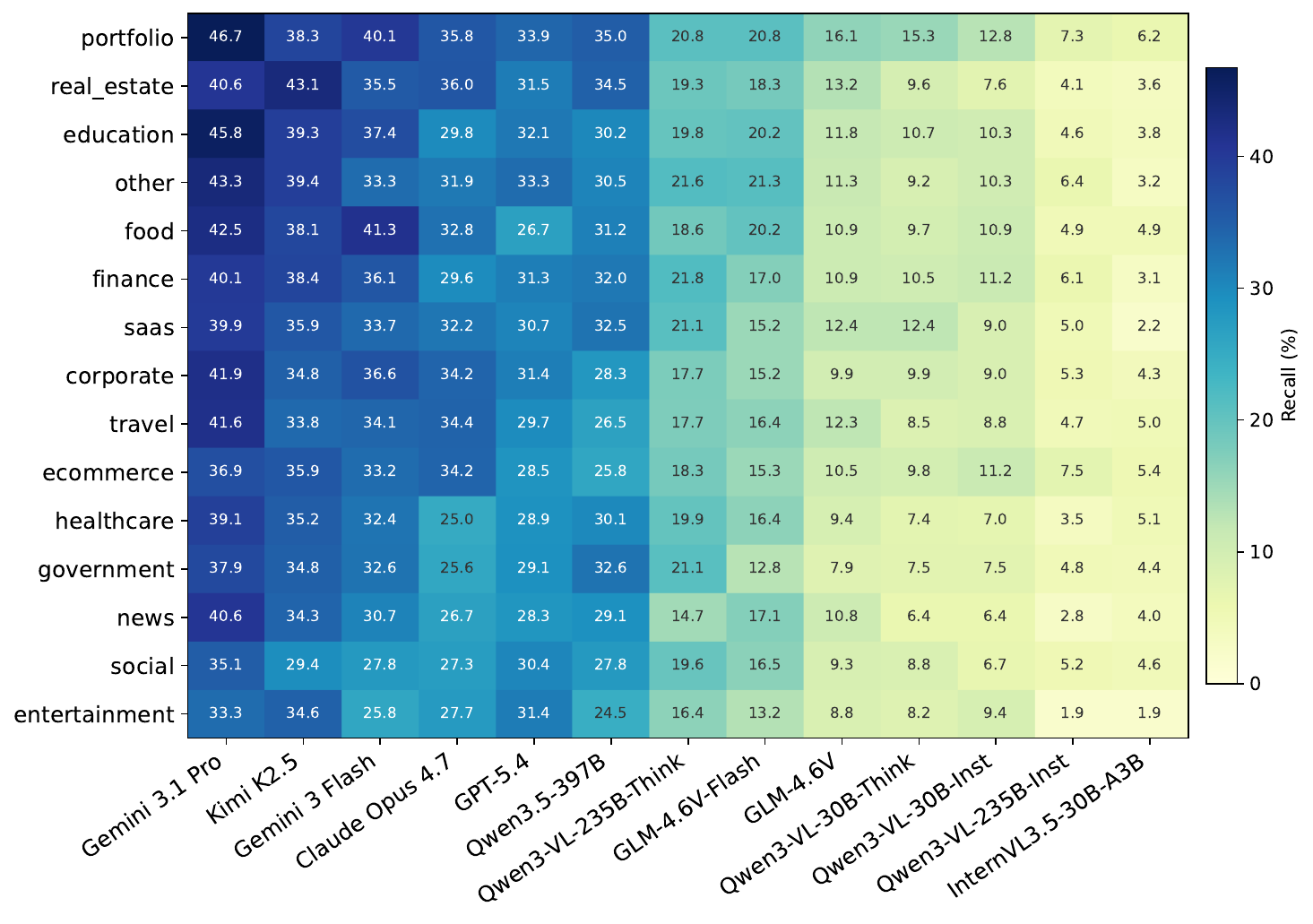}
    \phantomcaption
    \label{fig:per_domain_f1}
  \end{subfigure}
  \caption{\textbf{Recall heatmaps across 13 models.}
  Cells: has-diff Recall (\%). Columns: models, sorted by overall
  Recall (best at left). (a)~Rows: 13 CSS operators (300 has-diff
  pairs each). (b)~Rows: 15 source-domain labels
  (\S\ref{sec:construction}). The operator panel varies sharply by
  row while the domain panel is nearly flat: capability is
  property-specific, not domain-specific.}
  \label{fig:capability_heatmaps}
\end{figure}

\label{sec:property_failures}

Figure~\ref{fig:capability_heatmaps} compares recall along two axes:
CSS operators and source domains. The operator heatmap
(Figure~\ref{fig:heatmap}) shows large, structured differences across visual
properties. Some operators are consistently visible: \texttt{justify}, which
moves whole rows of text, reaches 87.0\% Recall with Gemini~3.1~Pro;
\texttt{text} substitutions can often be read through OCR, with
Claude~Opus~4.7 reaching 70.3\%; and \texttt{opacity} reaches 58.3\% with
Kimi~K2.5. In contrast, several operators remain difficult for nearly all
models: \texttt{gradient} reaches only 26.7\% at best, \texttt{line\_height}
has a median Recall of 4.0\%, and \texttt{rounded} has a median Recall of
13.3\%. The top-performing model also changes by operator: the gold top-1
outlines split across Gemini~3.1~Pro, Claude~Opus~4.7, Kimi~K2.5, and
Qwen3.5-VL-397B. Thus, an aggregate leader is not uniformly best across visual
properties.

The domain heatmap (Figure~\ref{fig:per_domain_f1}) shows the opposite
pattern. Recall varies much less across the 15 source-domain labels assigned
during construction. The easiest domain (\texttt{portfolio}, mean Recall
25.3\%) and the hardest (\texttt{entertainment}, mean Recall 18.2\%) differ by
only 7.1\,pp, while the best--worst model spread is 32.2\,pp. Model rankings
are also stable across domains: Gemini~3.1~Pro is top-1 on all 15 domain rows,
and the lowest-performing models remain at the bottom regardless of domain.
Together, Figure~\ref{fig:capability_heatmaps} shows that the capability gap is
property-specific rather than domain-specific: what changed matters more than
where it appeared.

\begin{figure}[t]
  \centering
  \begin{subfigure}{0.48\textwidth}
    \centering
    \includegraphics[width=\textwidth]{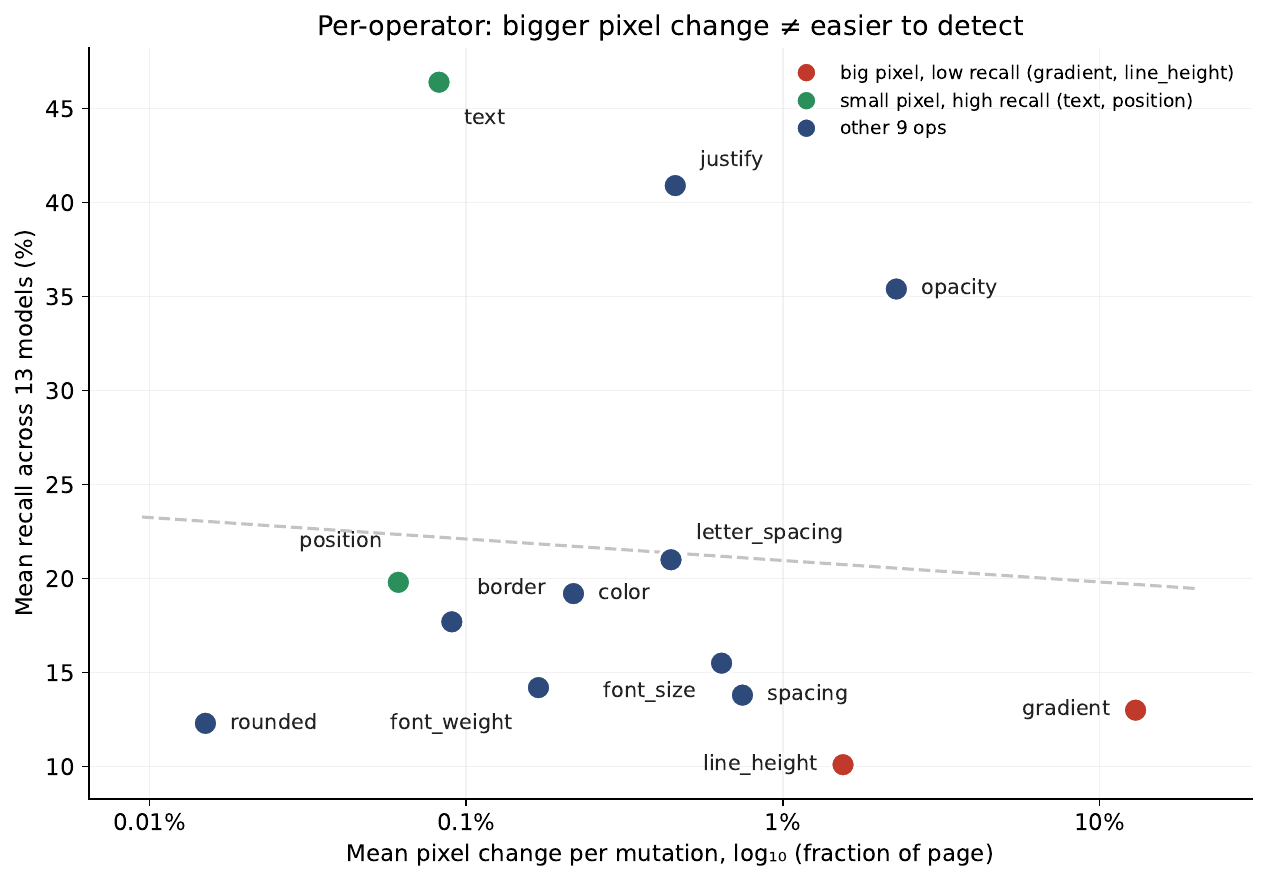}
    \caption{Pixel change vs.\ recall ($r = -0.08$).}
    \label{fig:pixel_vs_recall_a}
  \end{subfigure}
  \hfill
  \begin{subfigure}{0.48\textwidth}
    \centering
    \includegraphics[width=\textwidth]{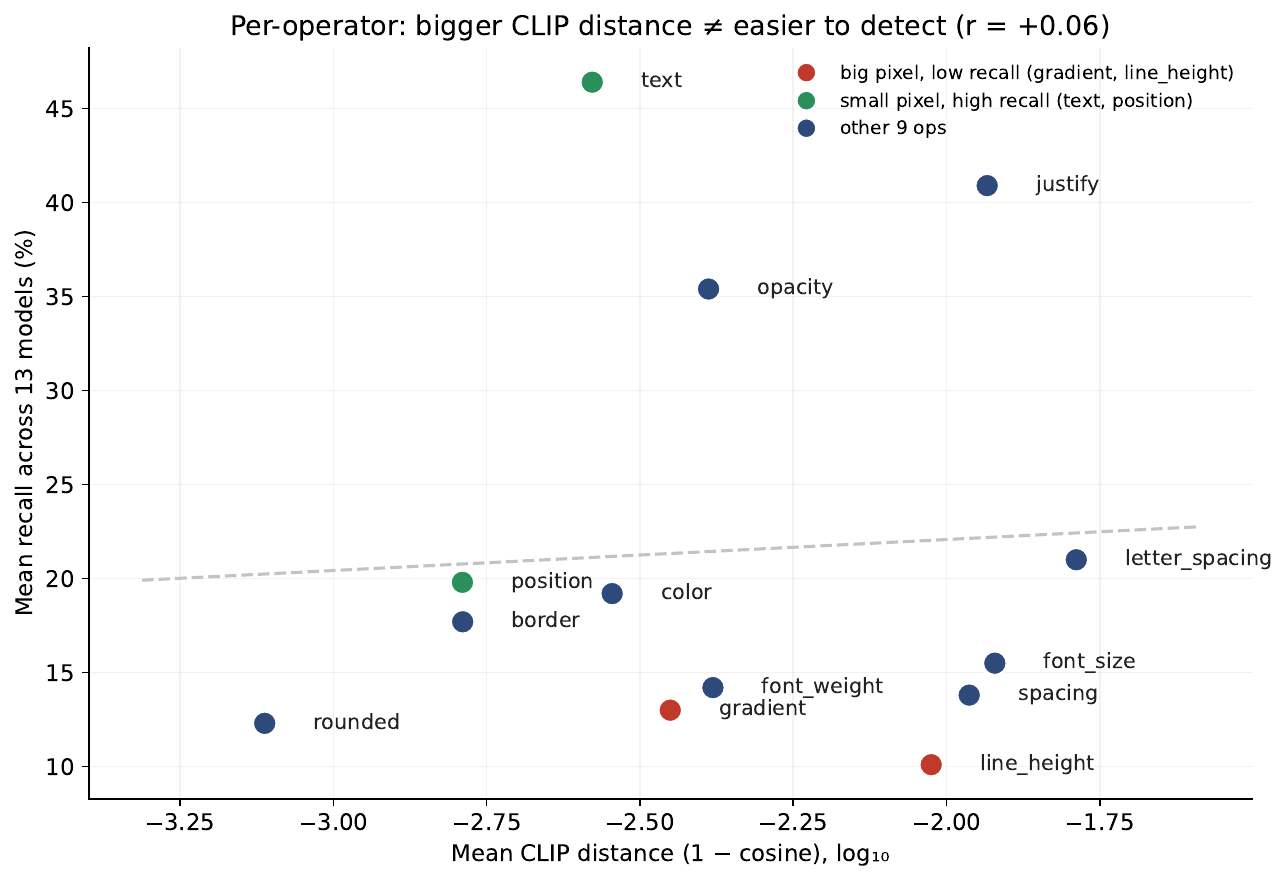}
    \caption{CLIP distance vs.\ recall ($r = +0.06$).}
    \label{fig:pixel_vs_recall_b}
  \end{subfigure}
  \caption{\textbf{Per-operator visual-signal magnitude vs.\ recall.}
  Each dot is one CSS operator; both axes use $\log_{10}$.
  Y: cross-13-model mean Recall on has-diff records (300 per operator,
  3{,}900 total). X (a): mean bbox-level pixel change
  per mutation (fraction of page). X (b): mean CLIP image-embedding
  distance ($1\!-\!\cos$).
  Both panels show a near-flat point cloud with effectively zero
  correlation. }
  \label{fig:pixel_vs_recall}
\end{figure}

\subsection{Visual Magnitude Does Not Explain Difficulty}
\label{sec:magnitude_analysis}

Figure~\ref{fig:pixel_vs_recall} tests whether operator difficulty can be
explained by the size of the visual signal. Across the 13 CSS operators,
bbox-level pixel change is essentially uncorrelated with mean Recall
($r=-0.08$, $r^2<1\%$; Figure~\ref{fig:pixel_vs_recall_a}). Operators such as
\texttt{text} and \texttt{position} occupy the low-pixel-change region but
achieve high Recall, while \texttt{gradient} and \texttt{line\_height} produce
larger pixel changes yet remain among the hardest operators. Thus, larger
pixel differences are not necessarily easier for VLMs to detect.

CLIP image-embedding distance does not explain difficulty either
($r=+0.06$, $r^2<1\%$; Figure~\ref{fig:pixel_vs_recall_b}). The CLIP distances
for all operators are compressed into a narrow range, suggesting that
caption-aligned image features are themselves insensitive to many
CSS-attribute-level mutations. The two views also fail in different ways:
\texttt{gradient} has large pixel change but relatively small CLIP distance,
while \texttt{letter\_spacing} has high CLIP distance because character
positions shift across a paragraph, even though model Recall remains modest.
These patterns suggest that DiffSpot difficulty is not governed by visual
magnitude alone, but by whether a model can perceive and name the changed
CSS-level visual property.

\subsection{Robustness to Judge Choice}
\label{sec:judge_robustness}

Because DiffSpot evaluates open-ended model responses, we verify that the
leaderboard is not an artifact of a single judge LLM. We re-score the full
13-model $\times$ 4{,}400-case grid with two additional judges
(\texttt{Kimi~K2.5} and \texttt{Qwen3.5-VL-397B}) using the same prompt and
visual-effect-equivalence rubric as \texttt{gpt-oss-120b}.
Across the three judge pairings, per-case agreement is high
(Cohen's $\kappa=0.92$--$0.94$), and the 13-model ranking is unchanged
(Kendall's $\tau=1.00$ for every pair, computed on the full grid).
The residual mean Accuracy differences ($\leq\!1.6$\,pp in any pair)
are small calibration shifts rather than leaderboard reshufflings. We therefore report main-table numbers under
\texttt{gpt-oss-120b}, the strictest judge; all comparative claims hold under
all three judges.

\begin{figure}[h]
  \centering
  \begin{subfigure}[t]{0.325\linewidth}
    \centering
    \includegraphics[width=\linewidth]{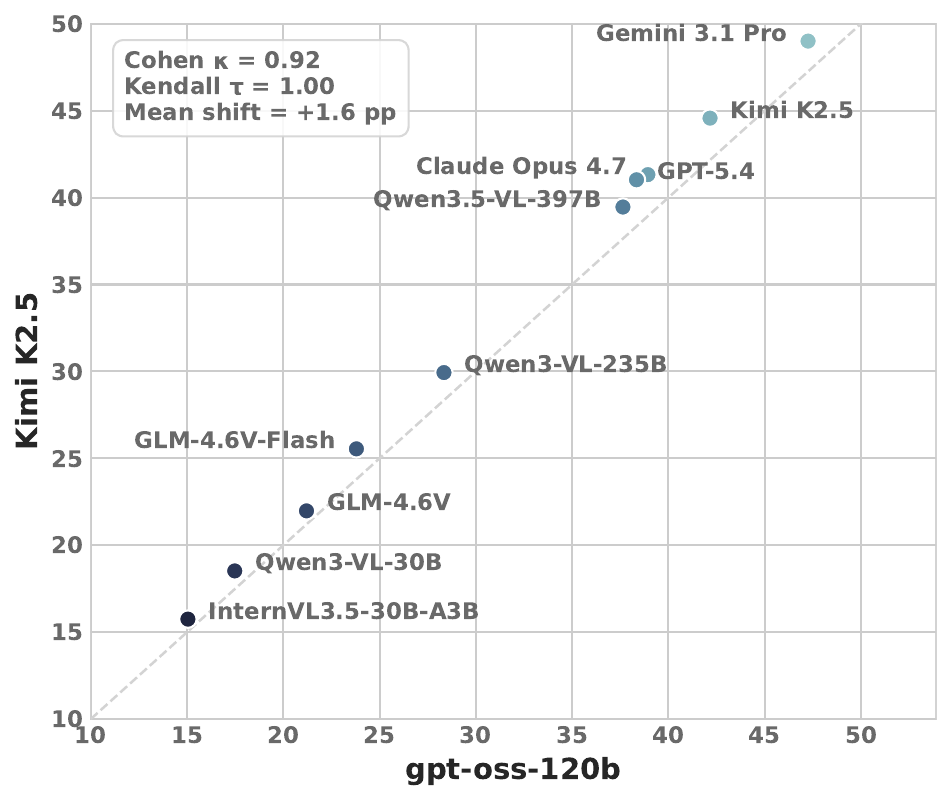}
    \caption{gpt-oss-120b vs.\ Kimi~K2.5}
    \label{fig:pairwise_a}
  \end{subfigure}\hfill
  \begin{subfigure}[t]{0.325\linewidth}
    \centering
    \includegraphics[width=\linewidth]{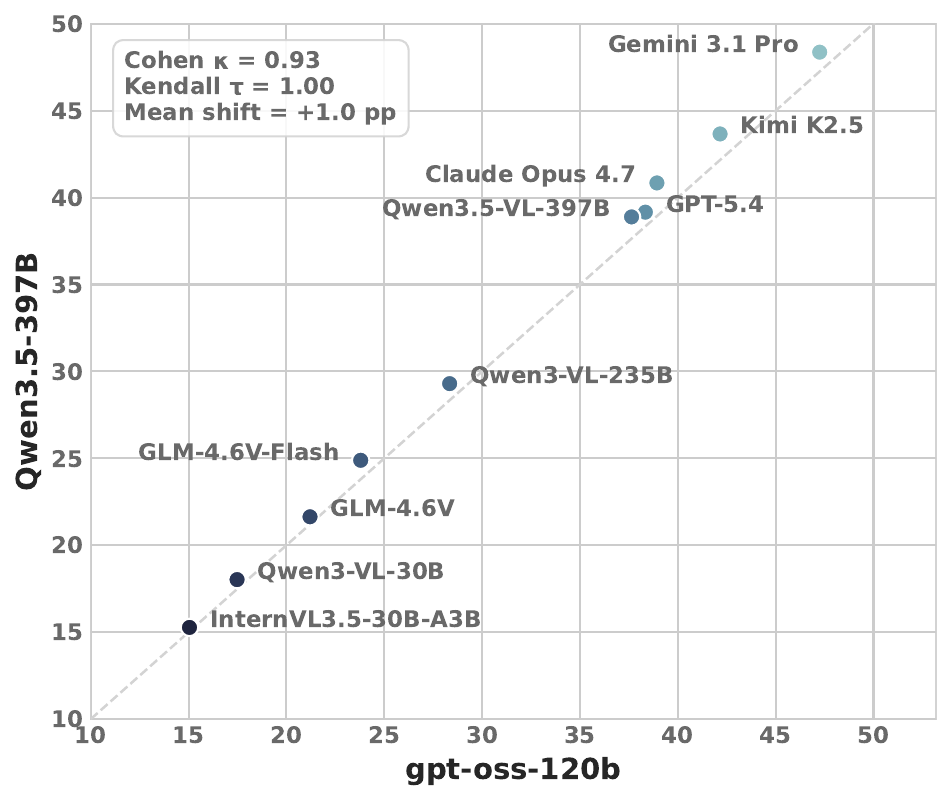}
    \caption{gpt-oss-120b vs.\ Qwen3.5-VL-397B}
    \label{fig:pairwise_b}
  \end{subfigure}\hfill
  \begin{subfigure}[t]{0.325\linewidth}
    \centering
    \includegraphics[width=\linewidth]{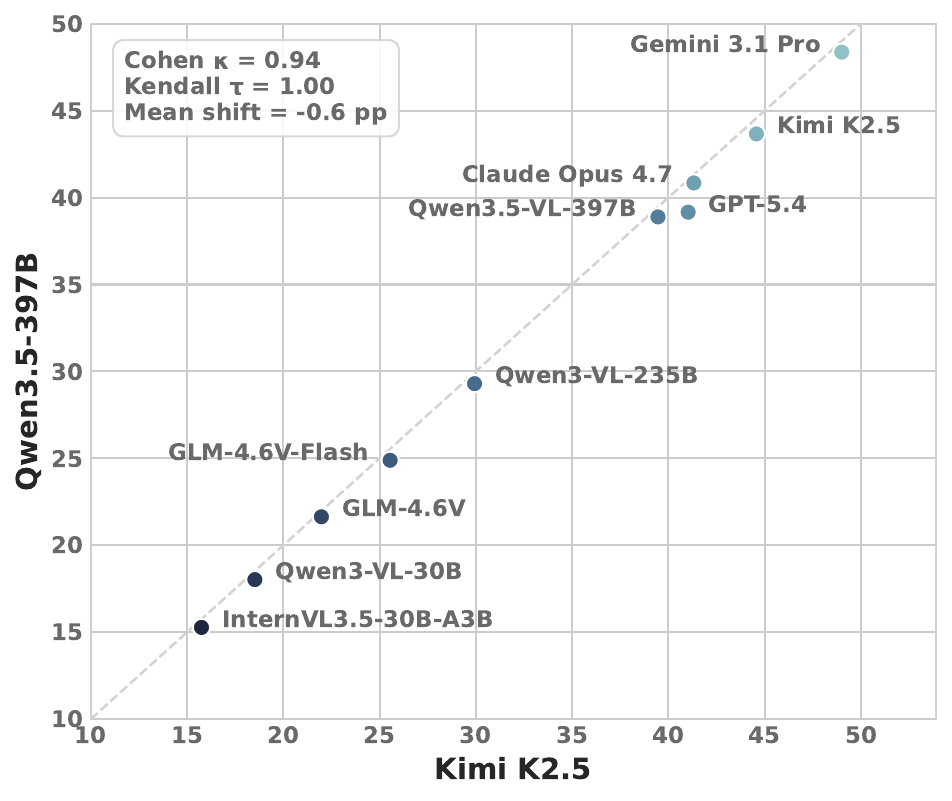}
    \caption{Kimi~K2.5 vs.\ Qwen3.5-VL-397B}
    \label{fig:pairwise_c}
  \end{subfigure}
  \caption{\textbf{Pairwise Accuracy agreement across three LLM judges.}
  Each dot is one VLM. Metrics are computed on all 13 models; the
  scatter omits two Qwen3-VL-Instruct variants and Gemini 3 Flash
  for visual clarity. Dashed line: $y=x$. Box: per-case Cohen's
  $\kappa$, Kendall's $\tau$ over the 13-model ranking, and mean
  Accuracy shift ($y$-axis judge vs.\ $x$-axis judge).}
  \label{fig:cross_judge_pairwise}
\end{figure}

\subsection{Realness Audit by Independent VLM Judges}
\label{sec:realness_audit}

We compare each original webpage screenshot against its
code-rendered counterpart by asking three VLM judges
(Qwen3.5-VL-397B, Kimi\,K2.5, Gemini~3 Flash) to rate every
image on a 1--5 realness scale in isolation, with no cue to
which image is which. Pairs are grouped by content richness into
\emph{rich}, \emph{standard}, and \emph{minimal}, and
Table~\ref{tab:fidelity} reports the paired mean difference
$\Delta = \mu(\text{orig}) - \mu(\text{rend})$ within each
group. All three judges rate originals and renderings within 0.3 points
on a 5-point scale, with renderings scoring marginally higher
across every group; the per-pair differences across judges are
essentially uncorrelated.

\begin{table}[H]
\centering
\caption{Paired realness $\Delta = \mu(\text{orig}) - \mu(\text{rend})$
on a 1--5 scale, stratified by visual style.}
\label{tab:fidelity}
\small
\begin{tabular*}{\textwidth}{@{\extracolsep{\fill}}lrrrr@{}}
\toprule
VLM judge       & rich    & standard & minimal & \textbf{Overall} \\
\midrule
Qwen3.5-VL-397B & $-0.15$ & $-0.22$  & $-0.34$ & $\mathbf{-0.23}$ \\
Kimi K2.5       & $-0.14$ & $-0.22$  & $-0.04$ & $\mathbf{-0.13}$ \\
Gemini 3 Flash  & $-0.16$ & $-0.34$  & $-0.40$ & $\mathbf{-0.30}$ \\
\bottomrule
\end{tabular*}
\end{table}

\section{Related Work}
\label{sec:related_work}

DiffSpot is a spot-the-difference benchmark for fine-grained visual
perception on rendered web interfaces. We position it against two
closely related lines of work: fine-grained visual perception
benchmarks (\S\ref{sec:rw_perception}) and visual-difference
benchmarks (\S\ref{sec:rw_diff}).

\subsection{Fine-Grained Visual Perception Benchmarks}
\label{sec:rw_perception}

A growing line of benchmarks probes whether VLMs perceive
fine-grained visual content beyond what high-level VQA requires.
Single-image perception probes reformat classic CV tasks as VQA,
MCQ, or yes/no
judgement~\citep{fu2024blink,tong2024eyes,li2024naturalbench,chen2024right,tong2024cambrian,kamath2023what};
image-pair and multi-image probes formulate comparative perception
as MCQ, yes/no, or paired
retrieval~\citep{comparebench,vlm2bench,statusbench,twin,vismin}; and a
parallel line studies object-presence, attribute, and relational
hallucination on single images via yes/no polling, open-ended VQA,
or caption-based CHAIR
analysis~\citep{pope2023li,hallusionbench2024guan,mmhal2023sun,amber2023wang}.
Other fine-grained perception benchmarks~\citep{mmdocbench,fgbmk,spec}
cover adjacent domains.

These benchmarks establish that VLMs can struggle with fine-grained
visual details, but they differ from DiffSpot in task form and
control. Most evaluate single-image perception or closed-form
comparative judgments, and each item is typically treated as a
single difficulty point. DiffSpot instead takes paired screenshots
as input, elicits an open-ended list of visual differences, and
places every has-diff item on a controlled
per-property~$\times$~per-magnitude grid.

\subsection{Visual Difference Benchmarks}
\label{sec:rw_diff}

The most direct comparison is to image-pair visual-difference
benchmarks. \textbf{OmniDiff}
~\citep{omnidiff} provides a large human-annotated image-difference
captioning dataset and evaluates generated captions with overlap
metrics such as BLEU, METEOR, ROUGE, CIDEr, and SPICE, which reward
paraphrastic similarity to a reference caption rather than
structural coverage of a change list. \textbf{VLM-SubtleBench}
~\citep{vlmsubtlebench} studies subtle paired-image differences
across multiple domains, but is primarily evaluated through
multiple-choice questions, with only a subset using captioning-style
evaluation.

A pre-VLM change-captioning lineage also studies visual differences,
typically pairing each dataset with a specialized captioning or
contrastive
model~\citep{spot_the_diff,ccexpert,btcchat,clevr_change,birds_to_words,vixen,brooks2023instructpix,imgdiff}.
Other image-pair benchmarks target adjacent capabilities, including
image-set differences, generalist diff-captioning, medical
difference VQA, and image-manipulation
description~\citep{visdiff,onediff,hu2023expert}. A
complementary line evaluates the generation side of image editing,
scoring whether a generative model can reproduce a target edit given
a known before/after
pair~\citep{magicbrush2023zhang,hqedit2024hui,emuedit2024sheynin};
DiffSpot instead evaluates the perception-side problem of identifying
what changed.

DiffSpot differs from these benchmarks along five axes. First,
\textbf{domain} focuses on rendered web interfaces rather than natural
images or generic image pairs, making it possible to specify visual
changes through the underlying HTML/CSS. Second, \textbf{ground truth}
is fully programmatic: each label is derived from a CSS mutation
record rather than human or LLM-generated annotation. Third,
\textbf{stratification} places each has-diff pair on a
per-property~$\times$~per-magnitude grid (13 operators $\times$ 3
difficulty tiers, 100 pairs per cell). Fourth, \textbf{evaluation}
elicits an open-ended diff list on every item and structurally
matches each response against the mutation record. Fifth,
\textbf{hallucination control} pairs the has-diff slice with 500
no-diff pairs, measuring false-positive difference reports under the
same protocol.

\section{Conclusion}
\label{sec:conclusion}

DiffSpot evaluates whether VLMs can identify fine-grained visual
differences on rendered web interfaces. By constructing pairs through
programmatic HTML/CSS mutation, it provides machine-readable ground truth,
controlled property--magnitude stratification, and open-ended diff-list
evaluation without caption-overlap or self-retrieval surrogates. Across 13 recent and frontier VLMs, the task remains far from solved:
even the strongest model identifies only 40.7\% of true visual changes,
and Hard-tier Recall stays below 23\% for every model. DiffSpot further
shows that failures are property-specific rather than domain-specific, that
pixel and CLIP magnitudes do not explain difficulty, and that no-diff pairs
expose a sensitivity--restraint trade-off. The benchmark, evaluation harness,
and self-contained HTML regeneration pipeline are released alongside the
paper.



\bibliographystyle{plainnat}
\bibliography{references}

\clearpage
\appendix


\section{Limitations and Broader Impacts}
\label{sec:limitations_impacts}

\paragraph{Limitations.}
\label{sec:limitations}
DiffSpot targets rendered web interfaces; extending the property-level
evaluation paradigm to mobile UI and desktop application screenshots is
left to future work. Source pages are predominantly English, so
multilingual and right-to-left layouts are a natural extension. The
no-diff metric counts whether a model fabricates any change; localizing
the specific false-positive claim is left to future work.

\paragraph{Broader impacts.}
\label{sec:broader_impacts}
DiffSpot supports research on fine-grained visual change detection for
web interfaces, with practical relevance to UI regression testing and
accessibility-oriented quality assurance. Better models and evaluation
tools in this area could reduce manual QA effort and make semantic UI
testing more accessible to smaller teams. As with other web automation
benchmarks, misuse is possible if similar techniques are applied to
monitor public webpages at scale. We mitigate data-release risks by
filtering candidate pages for personally identifiable information and
adult content, and by releasing only rendered screenshots of pages that
were publicly accessible at collection time.

\section{Visual-Reviewer Audit of the LLM Judge}
\label{app:judge}

Throughout the main paper we adjudicate VLM responses against the
ground-truth mutation using an LLM judge (\texttt{gpt-oss-120b}).
The cross-LLM-judge ranking-stability analysis is in
\S\ref{sec:judge_robustness}; this appendix documents a complementary
check that compares the LLM judge's verdicts against image-aware
adjudication on a random sample.

\paragraph{Setup.}
We draw a seed-42 random sample of $n{=}477$ unique judge verdicts.
Three independent volunteer reviewers---labelled \emph{Reviewer A},
\emph{Reviewer B}, and \emph{Reviewer C}---are each shown, for every
sampled case: (i)~the before/after screenshots, (ii)~the VLM's free-form
answer, (iii)~the ground-truth mutation description, and (iv)~the
judge's verdict together with its reasoning. Each reviewer returns an
audit verdict in $\{\texttt{correct},\texttt{wrong}\}$: \emph{correct}
means the reviewer, after looking at the images, agrees that the judge's
decision is consistent with what the VLM actually said versus what was
actually changed; \emph{wrong} means the reviewer believes the judge has
made an error visible from the image. Reviewers work independently; they
do not see each other's verdicts. The audit was a small voluntary expert
sanity check rather than a crowdsourcing or paid annotation study. The
reviewers were not used to construct the dataset or determine the
benchmark ground truth; they only inspected a random sample of LLM-judge
decisions to validate judge reliability.

\begin{table}[H]
\centering
\caption{Rate at which each independent visual reviewer agrees with the
judge's verdict, broken down by task type. The majority row
reports the fraction of cases on which at least two of three reviewers
agreed with the judge.}
\label{tab:judge-audit}
\small
\begin{tabular}{@{}lrrr@{}}
\toprule
Audit reviewer                 & has\_diff       & no\_diff        & \textbf{Overall} \\
\midrule
Reviewer A                     & $85.6\%$        & $98.3\%$        & $87.2\%$ \\
Reviewer B                     & $95.0\%$        & $98.3\%$        & $95.4\%$ \\
Reviewer C                     & $96.6\%$        & $100.0\%$       & $97.1\%$ \\
\midrule
\textbf{Majority} ($\geq$2/3)  & $\mathbf{96.9\%}$ & $\mathbf{98.3\%}$ & $\mathbf{97.1\%}$ \\
\bottomrule
\end{tabular}
\end{table}

\paragraph{Analysis.}
All three reviewers independently agree with the judge on at least
$87\%$ of the 477 audited cases; two of the three agree above $95\%$.
Taking the majority vote across reviewers as a best-available proxy for a
well-calibrated human gold standard, the judge's verdicts are
upheld on $97.1\%$ of cases overall---$96.9\%$ on \texttt{has\_diff} and
$98.3\%$ on \texttt{no\_diff}.
Inter-reviewer agreement is high: all three reviewers reach the same
verdict on $84.7\%$ of cases. Crucially, the fraction of cases that all
three reviewers unanimously flag as a judge error is only $5/477 = 1.0\%$,
which we take as the strongest empirically supported upper bound on the
judge's error rate on this sample. The remaining $\sim$14\% of cases
where reviewers split are overwhelmingly borderline judgments---partial
matches, paraphrases with ambiguous scope, or mutations on elements whose
visual manifestation is subtle---rather than clear-cut errors.

\paragraph{Takeaway.}
The LLM judge's decisions are almost entirely consistent with what
image-aware reviewers conclude on the same cases. Both the relative
model ranking and the absolute Accuracy numbers reported in the paper
are therefore robust to the choice of judging procedure. The 1\%
three-way-unanimous error rate provides a tight bound for readers
who wish to reason about residual noise in our evaluation.


\section{Accuracy--Hallucination Trade-off}
\label{app:tradeoff}

The aggregate Accuracy in Table~\ref{tab:main_table} combines two
behavioural axes---has-diff Recall and no-diff specificity---into a
single score. A two-dimensional view exposes failure modes that a
one-dimensional ranking hides.

\begin{figure}[h]
  \centering
  \includegraphics[width=0.8\textwidth]{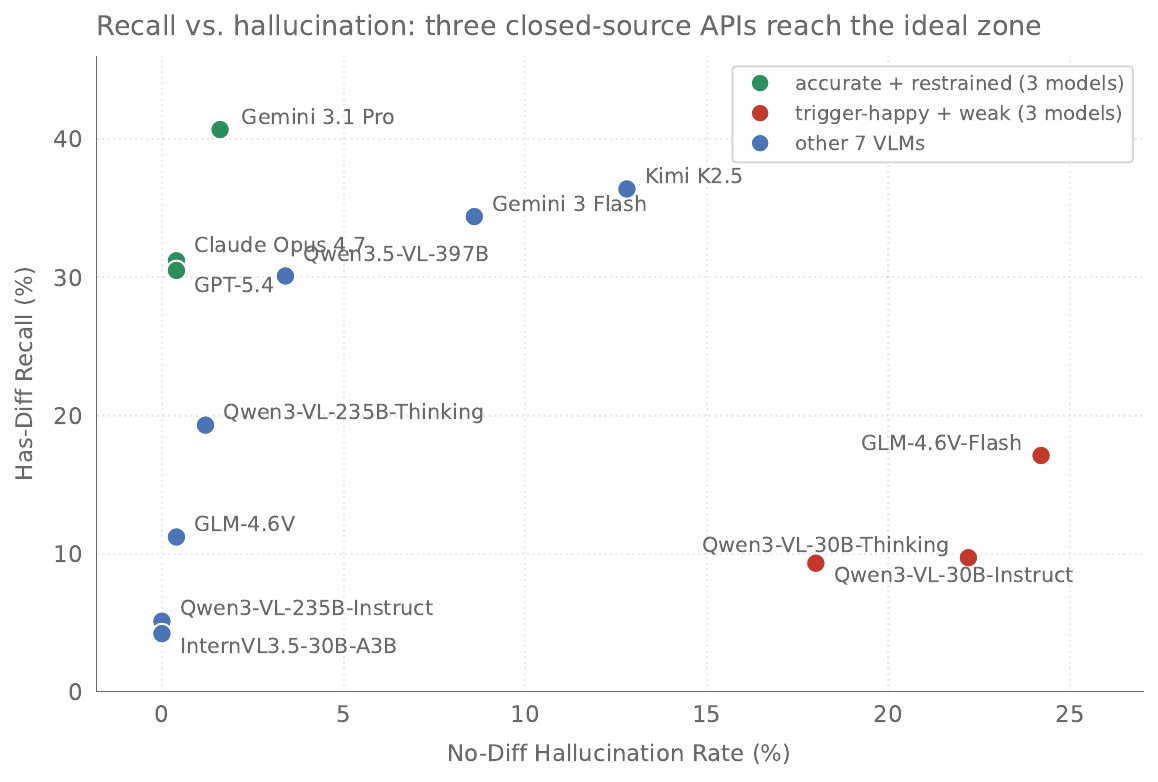}
  \caption{\textbf{Has-diff Recall vs.\ no-diff hallucination rate.}
  Each point is one of the 13 evaluated VLMs. The $y$-axis is overall
  Recall on the 3{,}900 has-diff pairs; the $x$-axis is the
  hallucination rate on 500 no-diff pairs (fraction reporting any
  change). \textcolor[HTML]{2A8F5A}{Green}: the three models in the
  ``accurate and restrained'' region (Gemini~3.1~Pro, Claude~Opus~4.7,
  and GPT-5.4). \textcolor[HTML]{C0392B}{Red}: three models that
  combine weak Recall with high hallucination.
  \textcolor[HTML]{4a74b4}{Blue}: the remaining seven. Models at
  halluc.\ $=$ 0.0\% (InternVL3.5-30B-A3B, Qwen3-VL-235B-Instruct)
  reach that position by producing near-empty outputs rather than by
  genuine selectivity.}
  \label{fig:f1_vs_halluc}
\end{figure}

Figure~\ref{fig:f1_vs_halluc} plots each model's overall has-diff
Recall against its no-diff hallucination rate---the fraction of 500
pixel-identical pairs on which the model still reports a change.
The 2D view exposes three distinct failure modes.
\textbf{(i) Three closed-source frontier APIs land in the
ideal zone} (Recall $\geq$ 30\%, hallucination $\leq$ 5\%):
Gemini~3.1~Pro (Recall 40.7\%, halluc.\ 1.6\%), Claude~Opus~4.7
(31.2\%, 0.4\%), and GPT-5.4 (30.5\%, 0.4\%) all achieve high recall
of real changes \emph{and} discipline when the input is unchanged.
No open-weight or privately-deployed model reaches this region.
\textbf{(ii) Three open-weight variants---GLM-4.6V-Flash
(Recall 17.1\%, halluc.\ 24.2\%), Qwen3-VL-30B-Thinking
(9.7\%, 22.2\%), and Qwen3-VL-30B-Instruct (9.3\%, 18.0\%)---are
simultaneously weak and trigger-happy}, the worst cost/benefit
combination; one in five to one in four of their no-diff outputs is
a fabrication.
\textbf{(iii) Models at halluc.\ $=$ 0.0\%} (InternVL3.5-30B-A3B and
Qwen3-VL-235B-Instruct) do \emph{not} exhibit principled selectivity:
their has-diff Recall of 4.2\% and 5.1\% shows that they report
almost nothing on any input, so they never hallucinate because they
never speak. Kimi~K2.5 and Gemini~3~Flash occupy the high-Recall
middle band but purchase their Recall with 9--13\% hallucination,
while Claude~Opus~4.7 (halluc.\ 0.4\%, Recall 31.2\%) and GPT-5.4
(0.4\%, 30.5\%) sit at the opposite extreme as the two
``conservative but competent'' operating points; Claude marginally
dominates GPT-5.4 on this plane (same hallucination rate, 0.7~pp
higher Recall). The Pareto frontier reduces to the segment from
Claude~Opus~4.7 to Gemini~3.1~Pro; every other model is dominated,
including GPT-5.4.

\section{Benchmark Size Justification}
\label{app:benchmark-size}

DiffSpot fixes 100 records per (operator $\times$ difficulty) cell,
giving 3{,}900 has-diff records (13 operators $\times$ 3 difficulties
$\times$ 100). We empirically verify that this size is sufficient to
produce a stable 13-model ranking using stratified sub-sampling.

\paragraph{Setup.}
For each $K \in \{10, 20, \ldots, 100\}$ records-per-cell, we draw
$200$ stratified random subsamples (each cell sampled independently),
recompute the 13-model has-diff Recall ranking on each subsample, and
measure Kendall's $\tau$ versus the full $K=100$ ranking. The
stratification fixes the operator/difficulty composition, so the only
varying factor is sample size. Top-1 model preservation and full
ranking exact-match rates are tracked alongside.

\begin{figure}[t]
  \centering
  \includegraphics[width=0.7\textwidth]{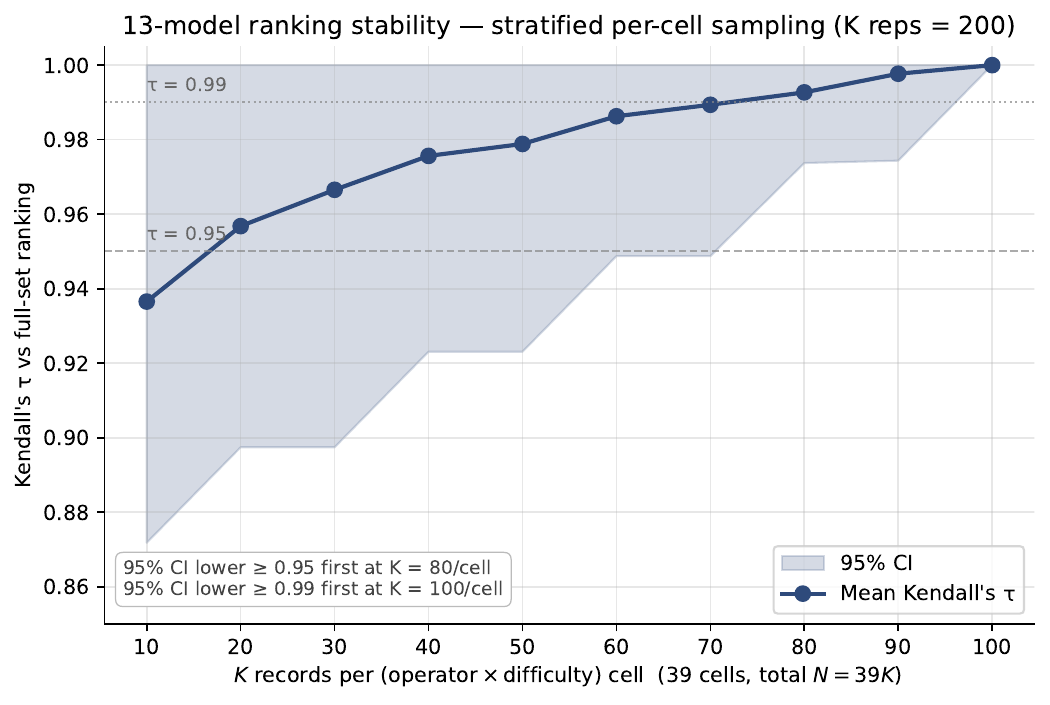}
  \caption{\textbf{Stratified per-cell ranking stability.} Mean
  Kendall's $\tau$ between the 13-model has-diff Recall ranking on a
  random subsample (drawing $K$ records per (operator $\times$
  difficulty) cell, 39 cells, $N = 39 K$) and the full ranking
  ($K = 100$, $N = 3{,}900$). Shaded band: 95\% CI over 200
  random subsamples per $K$. Dashed lines mark $\tau = 0.95$ and
  $\tau = 0.99$. The 95\% CI lower bound first reaches $0.95$ at
  $K = 80$ and $0.99$ at $K = 100$.}
  \label{fig:ranking_stability}
\end{figure}

\paragraph{Convergence.}
Figure~\ref{fig:ranking_stability} plots mean $\tau$ with the 95\%
confidence interval over the 200 reps. The lower bound of the 95\% CI
first reaches $\geq 0.95$ at $K = 80$ per cell ($N = 3{,}120$) and
$\geq 0.99$ at $K = 100$ per cell ($N = 3{,}900$, the paper setting).
The top-1 model (Gemini~3.1~Pro) is preserved in $\geq 99.5\%$ of
subsamples for every $K \geq 10$ and in $100\%$ for every
$K \geq 20$. The benchmark size we report sits exactly at the
operating point where ranking stability saturates: smaller $K$
produces visibly wider CI bands, while $K > 100$ would not be
reachable without enlarging the source pool.

\section{Mutation Mechanics}
\label{app:mutation-mechanics}

The mutator selects one of two mechanisms per operator:

\paragraph{(i)~Tailwind CSS class swap.}
For operators with a natural class taxonomy (\eg
\texttt{rounded-lg}$\to$\texttt{rounded-none},
\texttt{bg-blue-700}$\to$\texttt{bg-blue-100}), the mutator parses
the target element's \texttt{class} attribute with exact
\texttt{fullmatch} (no prefix-leak), skips responsive variants
(\eg \texttt{hover:}, \texttt{sm:}), and swaps in the new class.

\paragraph{(ii)~Inline \texttt{!important} style override.}
For \texttt{letter\_spacing} and \texttt{line\_height}---where
Tailwind's discrete scale does not span a range wide enough to
produce reliably non-trivial pixel deltas---the mutator appends
\texttt{style="letter-spacing: 0.12em !important"} (or analogous)
to the target element.

Both paths operate on static HTML before re-rendering.

\paragraph{Per-operator parameter ranges.}
Step-based operators (\texttt{rounded}, \texttt{color},
\texttt{font\_weight}, \texttt{font\_size}, \texttt{border},
\texttt{opacity}, \texttt{justify}, \texttt{position},
\texttt{spacing}, \texttt{gradient}) use a \texttt{max\_steps}
parameter bounding Tailwind-scale distance between before and after
values: Easy~$=$~3--5 steps, Medium~$=$~2, Hard~$=$~1.
Continuous-valued operators (\texttt{letter\_spacing},
\texttt{line\_height}) use em-offset magnitude: Easy~$=$~$\pm0.20$,
Medium~$=$~$\pm0.12$, Hard~$=$~$\pm0.06$. The \texttt{text}
operator uses character-substitution count (Easy~$=$~5+, Medium~$=$~2--4,
Hard~$=$~1).


\section{Construction Pipeline Details}
\label{app:construction-details}

\paragraph{CLIP-similarity gate threshold.}
The $\geq 0.70$ threshold was set via pilot inspection of $\sim$200
borderline pairs in the $[0.65, 0.75]$ band. Pairs above 0.70
preserved layout structure and font choices; pairs below began to
show layout drift.

\paragraph{Domain/style labeling.}
After PII / dynamic-tag / abnormal-length filters, an LLM labeler
(\texttt{gpt-oss-120b}~\citep{openai2025model} at temperature 0.3, with a
fixed prompt enumerating 15 domain categories and 4 visual-style categories)
assigns each page a domain and style tag. A capped-natural sampling policy
of at most 800 pages per domain bounds the largest domain to $\sim$7\% of
the corpus, yielding the multi-label enriched pool reported in
Figure~\ref{fig:pipeline} (Panel A).

\paragraph{Four-stage content filter (475 records removed).}
\begin{itemize}[leftmargin=*,itemsep=0pt,topsep=0pt]
\item \emph{Text-truncation} (96): the \texttt{text\_swap}
operator truncates element text to 50 characters for its diff
record; for long strings whose diff position falls beyond
character~50, the recorded \texttt{old} and \texttt{new} collapse
to the same prefix.
\item \emph{Bbox-failure} (190): mutations that push the target
out of the viewport make \texttt{getBoundingClientRect()} return a
null or clipped rectangle, breaking the outside-bbox locality
check.
\item \emph{Font-rendering tofu} (30): headless Chromium lacks
Noto Sans Korean / Thai / Devanagari, so pages containing those
scripts render as empty glyph boxes.
\item \emph{Viewport overflow} (168): if the target element
extends more than 30\% below the 800-pixel viewport bottom, the
visible portion of the change is too partial for fair comparison.
\end{itemize}
Filters are applied in union; after dedup of overlap, 475 records
are removed, leaving 20{,}629 \emph{filtered candidates}.

\paragraph{Cell-balanced sampling rationale.}
We pick $n=100$ per cell because the per-cell binomial standard
error at $p=0.5$ is $\approx 5$\,pp---tight enough for per-cell
reporting while staying within the supply of the scarcest
operator-tier cell after filtering.


\section{Prompts Used in Evaluation}
\label{app:prompts}

For full reproducibility we reproduce the prompts used for both the
VLM evaluation step (\S\ref{app:vlm-prompt}) and the LLM judge step
(\S\ref{app:judge-prompt}). Verbatim text is shown exactly as passed
to the models at inference time.

\subsection{VLM Prompt}
\label{app:vlm-prompt}

Every evaluated VLM receives the two rendered screenshots (before /
after) together with the user prompt shown below. The wording uses
the permissive ``\textit{a change may have been made}'' form rather
than asserting that a change has occurred. Asserting a change primes
the model on no-diff inputs and inflates the hallucination rate; the
permissive phrasing leaves both has-diff and no-diff records as
first-class cases.

\begin{promptbox}{VLM User Prompt}
Look at these two webpage screenshots carefully (Image 1 = before,
Image 2 = after).
A change may have been made between the two screenshots.

Your task: Find and describe ALL visual differences you can spot.
- Be specific about the location (e.g., "the heading at the top",
  "the button in the sidebar")
- Describe what changed and how (e.g., "the text became bolder",
  "the spacing increased")
- Don't skip any difference, no matter how small
- IMPORTANT: Only report differences you can actually SEE in the images.
  Do NOT guess or make up differences.
- If you cannot find any difference, say "No visible differences found."

List each difference on a separate line starting with "- ".
\end{promptbox}

\subsection{Judge Prompt}
\label{app:judge-prompt}

The judge prompt is a templated document composed of (i)~a fixed base
template injecting the ground-truth mutation, the VLM's free-form
answer, and general element-matching rules; and (ii)~one
operator-specific rule block injected at \texttt{\{operator\_rule\}}
based on the ground-truth mutation type (see
\S\ref{app:operator-rules}). For no-diff cases the operator slot is
replaced by a no-diff rule. Per-call prompt length is
${\sim}$1{,}100--1{,}350 tokens.

\begin{promptbox}{Judge Prompt: Base Template}
You are evaluating whether a Vision-Language Model (VLM) response
describes a specific webpage change. You will compare the VLM's text
against either a single ground-truth change, or no change at all.

## Ground Truth Change
{gt_json}

## Ground Truth Description (reference only)
{gt_answer}

## VLM Response
The VLM was shown both screenshots and asked to find all visual
differences. Here is the VLM's response:
---
{vlm_response}
---

## General element matching

- `header` ~ "top nav" / "top bar"; `hero section` ~ "banner";
  `footer` ~ "bottom section"; `button` ~ "CTA"; `card` ~ "tile" /
  "panel"; `heading` ~ "headline" / "h1/h2".
- Container <-> child: when GT says "section X changed" and VLM
  describes children of X all showing the change -> match (effects
  like opacity propagate to children).
- Text matched by quoted content: when VLM quotes the SAME before/after
  text as GT -- match, regardless of how VLM labels the section.
- Quoted GT text is a prefix: the quoted text inside the GT description
  identifies the element by its opening characters, not the full
  string. A VLM quote that starts with the same characters and
  continues further refers to the same element.
- Distinctive text token overrides location labeling: If both GT and
  VLM reference a distinctive text token (button label, menu item,
  author name, etc.), they refer to the SAME element even if VLM
  labels its surrounding region differently from GT.

## PRINCIPLE OVER KEYWORDS

The operator-specific rule below has an ESSENCE (the physical/visual
reality) and ACCEPT/REJECT criteria. Apply the principle, not keyword
matching. The ACCEPT examples are illustrative -- any VLM phrasing
that semantically matches the principle should be accepted, even if
wording differs from examples. REJECT only when the description
genuinely contradicts the physical effect.

## Operator-specific Rule

{operator_rule}

## Rules of thumb

- Direction consistency required for all non-opacity operators.
- Color-name precision NOT required if direction matches.
- Extra claims by VLM = hallucinations, NOT verdict-wrong: if VLM
  correctly paraphrases the GT change AND adds extra unrelated claims,
  verdict stays correct; the extras go to `hallucinations`.
- Conclude "wrong" ONLY after applying the operator-specific principle
  and confirming no semantic match.

## Your Task

Step 1: Reasoning. Write out:
  1. What does the GT change describe (element, property, direction)?
  2. Which parts of the VLM response could refer to this change?
     Quote the relevant text.
  3. Apply the operator-specific principle above. Determine if the
     VLM's description matches the ESSENCE or ACCEPT criteria. If no
     match, check whether it falls under REJECT or is simply unrelated.

Step 2: Verdict. Assign one of:
  - correct: VLM identifies the GT element specifically AND describes
    a change matching the operator's ACCEPT criteria.
  - wrong: VLM did not mention the element specifically; OR mentioned
    the element but did not claim any change; OR described a different
    property / reversed direction failing the operator's REJECT
    condition.

## Hallucinations

After the verdict, list every VLM-reported difference that does NOT
correspond to the GT change. Each distinct extra claim is a separate
entry. If GT is null, every specific VLM-reported difference is a
hallucination.

## Output

Return ONLY valid JSON (no markdown fences, no commentary):

{
  "reasoning": "step 1 analysis with operator-specific principle check",
  "gt_verdict": "correct" | "wrong" | null,
  "vlm_match": "quote from VLM that matches the GT change, or empty",
  "hallucinations": ["each distinct hallucinated difference"],
  "summary": {"correct": 0, "wrong": 0, "hallucinated": 0}
}

If GT is null, set `gt_verdict` to null, `vlm_match` to empty,
correct/wrong to 0.
\end{promptbox}

\begin{promptbox}{Judge Prompt: No-diff Rule (injected when GT is null)}
**This case: NO DIFFERENCE** (the two screenshots are identical)

There is no GT change. The verdict is null. Any specific visual
difference reported by the VLM is a hallucination.

ACCEPT (verdict null with empty halluc list):
- VLM correctly says no differences ("no visible differences found",
  "the two images appear identical", etc.)

REJECT (verdict null but each VLM claim becomes a hallucination):
- VLM reports any specific change (color, position, spacing, text,
  etc.) -- list each as a separate hallucination
\end{promptbox}

\subsection{Per-Operator Rules}
\label{app:operator-rules}

For has-diff cases the judge receives one of 13 operator-specific rule
blocks, selected by the ground-truth mutation type. Every rule follows
the same four-part structure: (1)~a one-sentence PRINCIPLE describing
the physical/visual signature of the operator; (2)~an ACCEPT section
listing paraphrases and anchor phrasings; (3)~a REJECT section
specifying what counts as a genuine contradiction; (4)~confirmed
anchor phrasings observed across models. Table~\ref{tab:op-principles}
summarises the thirteen principles.

\begin{table}[h]
\centering
\caption{One-line principle for each of the 13 per-operator judge
rules. The full ACCEPT/REJECT text is released with the evaluation
code; two illustrative full rules (\texttt{opacity},
\texttt{spacing}) are reproduced below.}
\label{tab:op-principles}
\small
\begin{tabular}{@{}p{0.17\linewidth}p{0.78\linewidth}@{}}
\toprule
Operator & Principle \\
\midrule
\texttt{opacity}        & Opacity decrease blends the element's colour toward the background; direction (lighter/darker) depends on background luminance. \\
\texttt{position}       & Element moves $N$ pixels in the GT direction; adjacent layout (gaps, overlaps, line wraps, separators) changes as a consequence. \\
\texttt{spacing}        & Container padding / gap changes; contents shift away from the padded edge when padding increases, toward it when it decreases. \\
\texttt{justify}        & Container's child items redistribute along the main axis; any individual child's shift is valid evidence of the redistribution. \\
\texttt{letter\_spacing} & Gap between characters changes; text (and containing button/label) becomes wider or narrower. \\
\texttt{font\_weight}   & Character stroke thickness changes; bolder = thicker, lighter = thinner. \\
\texttt{font\_size}     & Character size changes; text block occupies more/less space and may wrap to more/fewer lines. \\
\texttt{line\_height}   & Vertical spacing between lines of a paragraph changes (looser / tighter). \\
\texttt{color}          & Colour moves to a darker or lighter shade within the same hue family; colour-name precision is tolerant. \\
\texttt{border}         & Colour of the visible outline changes; ``border'', ``line'', ``outline'' are treated as synonyms. \\
\texttt{rounded}        & Corner shape changes (sharp $\leftrightarrow$ rounded/pill/circular). \\
\texttt{gradient}       & Axis of colour transition changes (horizontal $\leftrightarrow$ vertical $\leftrightarrow$ diagonal, or reversal of the same axis); positional colours rearrange. \\
\texttt{text}           & Specific characters / words / punctuation are altered; the exact before$\to$after substitution is the identifier, region labelling secondary. \\
\bottomrule
\end{tabular}
\end{table}

\begin{promptbox}{Judge Prompt: Operator Rule for \texttt{opacity}}
**This case: OPACITY decrease** (e.g., 100% -> 50%/25%)

PRINCIPLE: When opacity decreases, the element's color blends toward
the background color (semi-transparent overlay). Direction depends on
bg: dark element on light bg -> appears LIGHTER; light element on dark
bg -> appears DARKER. Both directions valid as long as the new color
is closer to surrounding bg.

Key paraphrase equivalence: The visual signature of opacity reduction
IS a color shift on the same element. When VLM correctly identifies
the GT element AND describes any perceptible color change on it
(shade shift, "became lighter", "became faded", "changed to lighter
blue", "pale version of the original color"), this is a valid
description -- even if VLM does not use the word "opacity" or
"transparent".

CRITICAL -- Judge cannot see images, so cannot verify bg color directly.
For opacity DECREASE, ACCEPT EITHER direction (lighter OR darker) when:
- VLM describes consistent direction across multiple sub-elements
  (e.g., "text + button + placeholder all became darker") -> strong
  evidence of uniform opacity blend toward bg
- VLM uses opacity-language ("faded / ghosted / transparent /
  semi-transparent") in EITHER direction -> ACCEPT regardless of
  light/dark bg assumption
- Direction-only descriptions ("became darker" / "became lighter") on
  the GT element -> ACCEPT (Judge cannot verify which direction is
  "right" without bg)

REJECT direction ONLY when:
- VLM says "more vivid / more saturated / brighter and bolder /
  sharper contrast" -> clearly OPPOSITE effect (increased opacity)
- VLM describes hue shift unrelated to bg blend (e.g., "red -> blue"
  on a white bg)
\end{promptbox}

\begin{promptbox}{Judge Prompt: Operator Rule for \texttt{spacing}}
**This case: SPACING/PADDING change** (container padding/margin/gap
increased or decreased)

PRINCIPLE: Container's inner space changes. Contents physically shift
AWAY from the padded edge when padding INCREASES, and TOWARD the edge
when padding DECREASES. For gap changes (between sibling items), gap
visibly grows/shrinks.

No CSS terminology required: VLMs see visual consequences, not CSS
properties. ACCEPT descriptions of visual consequences WITHOUT
requiring the VLM to name "padding" / "margin" / "gap" / any CSS
keyword. Valid phrasings include:
- "the box / container became wider / narrower / taller / shorter"
- "there is more / less empty space between X and Y"
- "X is closer to / further from Y"
- "the content shifted inward / outward / got pushed to the right"
- "elements are more spread out / more clustered / closer together"
- "text has more / less room around it"

ACCEPT any description matching this principle:
- Content shift (consistent direction): left padding UP -> "contents
  pushed right"; vertical padding UP -> "X moved downward" / "more
  space above"
- Container resize: "box visibly wider/taller" (for UP) or
  "narrower/shorter / more compact" (for DOWN)
- Empty-space descriptions: "there is more empty space between the
  left border of the box and the content" -- ACCEPT even without
  saying "padding"
- Internal element changes: "input fields are wider" (when container
  padding DOWN frees internal space)
- Text wrap consequence: "text wraps to more lines" / "fewer lines"
- Gap changes: "horizontal/vertical spacing between X and Y
  decreased/increased"

REJECT only when:
- Direction genuinely contradicts (padding UP described as "content
  shifted toward edge / less space" is opposite)
- VLM describes a completely unrelated element that is not in or
  adjacent to the GT container
\end{promptbox}

The remaining eleven per-operator rules follow the same
ESSENCE / ACCEPT / REJECT / anchor-phrasings structure, with
rule-specific examples tuned to each operator's visual signature. The
full text is released with the evaluation code
(\texttt{scripts/09\_judge\_single\_model.py}, function
\texttt{get\_operator\_rule}).

\section{Compute Resources}
\label{app:compute}

The full evaluation grid comprises $13 \times 4{,}400 = 57{,}200$
open-ended VLM generations on the benchmark, scored by three
independent judge LLMs for a total of $\sim 1.7 \times 10^{5}$
judge verdicts. Closed-source frontier models (Anthropic, OpenAI,
Google families) are queried through their public APIs at default
decoding temperatures. Open-weight VLMs and the judge LLMs are
served on NVIDIA H20 GPU pods. We do not report wall-clock figures
because end-to-end runtime is dominated by external API rate limits
rather than local compute, but the H20 footprint is sized to
comfortably hold the largest evaluated open-weight model (a
235B-parameter MoE) alongside the \texttt{gpt-oss-120b} judge, and
the full grid was reproduced more than once during pipeline
development.

\end{document}